\documentclass{article}

\usepackage[final]{neurips_2020}

\usepackage[utf8]{inputenc} 
\usepackage[T1]{fontenc}    
\usepackage{hyperref}       
\usepackage{url}            
\usepackage{booktabs}       
\usepackage{amsfonts}       
\usepackage{nicefrac}       
\usepackage{microtype}      
\usepackage{multicol}
\usepackage{etoolbox,refcount}
\usepackage{graphicx}
\usepackage{biblatex}
\usepackage{subfig}
\usepackage{textcomp}

\title{Unsupervised learning for economic risk evaluation in the context of Covid-19 pandemic}

\author{
  Santiago Cortés\\
  Factored.ai\\
  Medellin, Antioquia  \\
  \texttt{santiago.cortes@factored.ai} \\
   \And
 
  Yullys María Quintero\\
  Department of Computer Science\\
  EAFIT university\\
  Medellin, Antioquia  \\
  \texttt{ymquinterm@eafit.edu.co} \\
}

\begin{document}

\maketitle

\begin{abstract}
  Justifying draconian measures during the Covid-19 pandemic was difficult not only because of the restriction of individual rights, but also because of its economic impact. The objective of this work is to present a machine learning approach to identify regions that should implement similar health policies. For that end, we successfully developed a system that gives a notion of economic impact given the prediction of new incidental cases through unsupervised learning and time series forecasting. This system was built taking into account computational restrictions and low maintenance requirements in order to improve the system's resilience. Finally this system was deployed as part of a web application for simulation and data analysis of COVID-19, in Colombia, available at (\url{https://covid19.dis.eafit.edu.co}).     
  
\end{abstract}

\section{Introduction}

The Covid-19 pandemic brought with it both health and economic problems for public administrations. Lockdowns \cite{lau2020positive}, increased Government spending \cite{dell2020economic}, and compulsory mask usage \cite{feng2020rational} were among the measurements taken by governments. Some of these measures were claimed as necessary but were unpopular among the citizens, mainly because of their economic impact, as well as the restriction of individual liberties. Public policies during the Covid-19 outbreak should be based on holistic approaches that consider health, economical and social variables in a given territory. There have been some  unsupervised approaches to establish the economic impact of Covid-19 pandemic(\cite{carrillo2020using} \cite{rahman2020data}), but those studies are focused exclusively on health variables.

In this work we propose a system that generates clusters among Colombian regions based on Covid-19 new incidences forecasts, health, geographic, demographic and economic variables (see Table \ref{table1}). Hence, such clusters generate geographical regions with similar impact due to the pandemic in several dimensions. This delivers public administrators a source to identify vulnerable population and identifying the impact of policies in different regions, thus providing an extra input for the generation and justification of public policies among uncertainty to contain the pandemic at a regional level.

\section{Design}

The system proposed consists of two parts. On the one hand, a forecasting model that predicts the number of cases for each of the next 7 days in all 33 Colombian departments. On the other hand, a clustering process that, with the help of some dimensionality reduction techniques, generates clusters associated to each department. 

The forecast of new incidences is included in a vector representation that also contains the variables shown in Table \ref{table1} for each department. Those variables are static in time, as it is assumed that they do not change significantly during each week of forecast. The demographic and economic variables were extracted from the official sources of the National Administrative Department of Statistics (DANE) and the National Planning Department (DNP). The information of the geographical variables was obtained from the official page of Colombia’s Institute of Hydrology, Meteorology and Environmental Studies (IDEAM). The health variables are available on the official web sites of the National Institute of Health (INS), and on that of the Ministry of Health and Social Protection.
\begin{table}[!htbp]
\caption{Recollected variables}
\label{table1}
 \centering
\resizebox{\textwidth}{!}{%
\begin{tabular}{lllll}
\multicolumn{5}{c}{Gathered variables}                                                                                                              \\  \hline
altitude                  & population between 15 and 24 years & Child labour     & Total population                & population with Diabetes      \\
precipitation             & population over 65 years           & Dependency ratio & Life expectancy                 & Deaths by chronic diseases    \\
temperature               & population density                 & Informal economy & Deaths by digestive diseases    & Deaths by acute diseases      \\
humidity                  & women population                   & illiteracy       & Deaths by respiratory illness   & Deaths by endocrine disorders \\
population under 15 years & Multidimensional Poverty Index     & school dropout   & Deaths by cardiac complications & Death by maligne Neoplasm     \\  \hline
\end{tabular}}
\end{table}

These embeddings make a vector representation of each political subdivision of interest that contains, the pandemic’s progression, economical, healthcare, geographical and demographic data. Then, the embeddings are clustered (after applying a dimensionality reduction technique) and such clusters are the system’s final output.

\begin{figure}[!htbp]
  \centering
\includegraphics[scale=0.2]{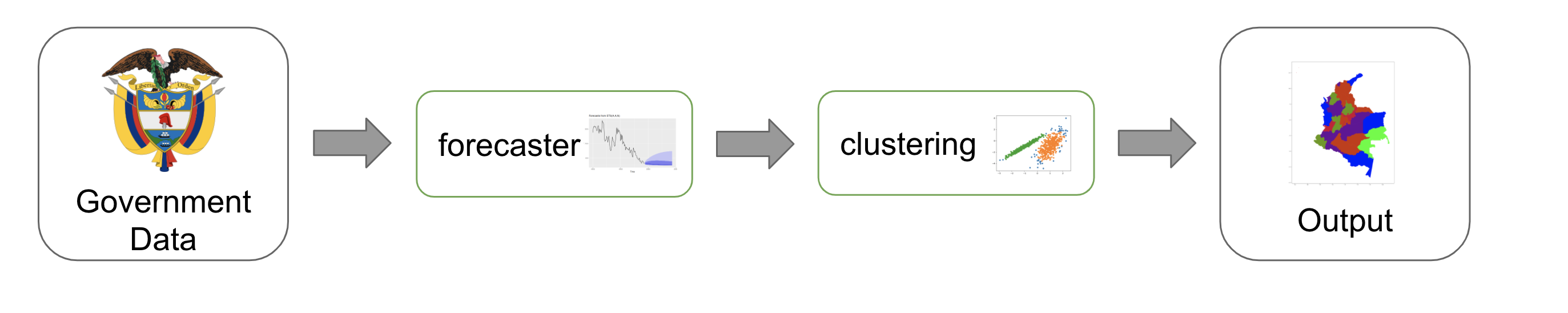}
 \caption{System's architecture.}
\end{figure}

\section{Implementation}
\subsection{Forecaster}

The chosen forecaster model is a neural net that only uses the new Covid-19 cases as external variables (this is due to the scarcity of APIs with updated public data). The series of new positive cases by department have several peaks, explained maybe by delays in testing. To overcome this problem, three features were included in the model. First, the input data is normalized by taking the logarithm of the raw data. Then, an exponential moving average (equation 1), with $\alpha=0.1$, is computed for all the new incidences series and used as feature(as is suggested in \cite{smyl2020hybrid}). Finally, the day of the week was incorporated as an input using an embedding layer.

\begin{equation}
    v_{t} =\alpha x_{t-1} - (1 - \alpha)v_t
\end{equation}
 
Encoder-decoder architectures have been satisfactorily used for time series forecasting \cite{sutskever2014sequence}. In particular, 2014 deep Mind’s wave net \cite{oord2016wavenet} proposed a way to stack convolutional layers to extract features from sequence data, and keep the number of parameters in the model low. This last property reduces the amount of computational power required to train and deploy the system. This is the reason why, the used architecture consists of an encoder that is made of a series of blocks, each one composed of a dilated convolution with 128 filters of size 2 and causal padding, a Dense layer followed by batch normalization, and finally, a ReLu activation \cite{jin2015deep}. There were 6 of these blocks with 1,2,3,4,5 and 6 as dilation rates, respectively. The decoder takes as inputs the encoder output and one dimensional embeddings for the day of the week in the forecasted window, and then passes these inputs through to dense layers with ReLu activations. The predictions are made by a dense layer with the size of the forecasted windows as its number of units.

\begin{figure}[!htbp]
  \centering
  \subfloat[Forecasting net architecture]{\includegraphics[scale=0.3]{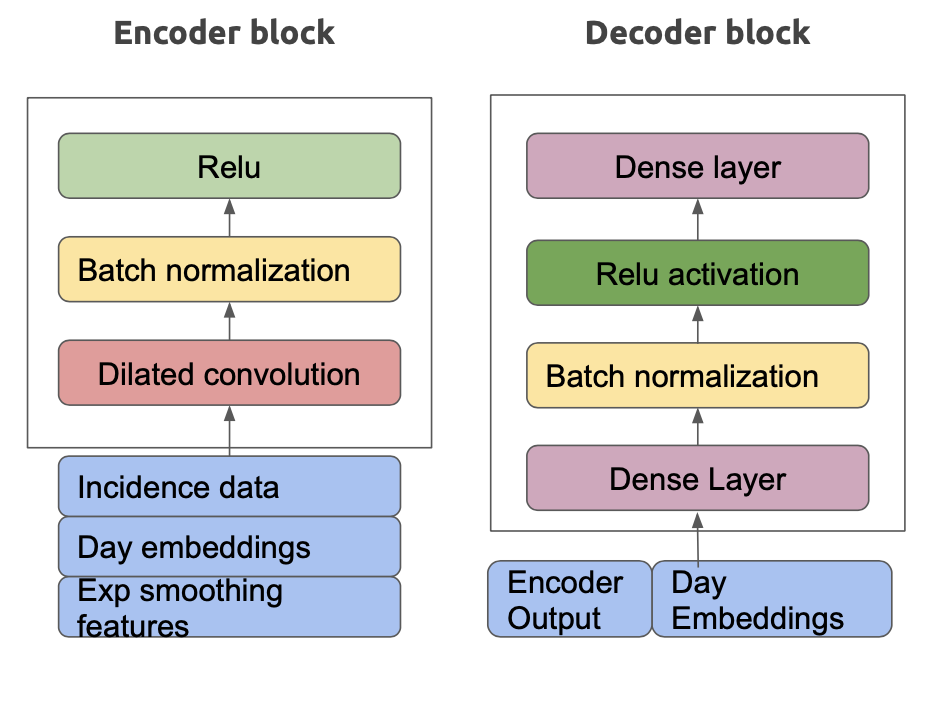}\label{f1}}
  \hspace{1em}
  \subfloat[Symmetric autoencoder architecture.]{\includegraphics[scale=0.5]{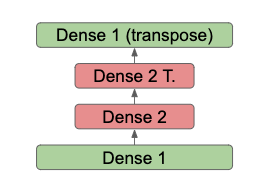}\label{f2}}
  \caption{Deep learning architectures used in this work.}
  \label{fig:architectures}
\end{figure}

The net is trained using the quantile loss in order to predict confidence intervals. The 5, 90 and 50 quantiles are predicted, and the last one is taken as the prediction. Before computing the loss, the net output is re-scaled using the last value of the exponential smoothing in the training window:
\begin{equation}
\hat{y}_{t+1,..t+h} = exp(NN(x))*I_t
\end{equation}

\subsection{Clustering}
Once the inference is made, the forecasted mean of the predicted window is added to a vector that contains all the other  gathered variables. Those vectors are normalized and then, in order to avoid the curse of dimensionality \cite{friedman2001elements}, a dimensionality reduction is done before applying the k-means or k-medoids algorithm.

Three dimensionality reduction techniques  were used. The first one, was a PCA that captures 90 percent of the original variance. Then, we used genetic algorithms for feature selection as they can be used to improve silhouette scores \cite{lleti2004selecting}. Lastly, we tried two simple auto-encoders with the same architectures. In the symmetric one the weights between layers are shared, thus reducing the number of weights as pictured in Figure \ref{fig:architectures}. Both auto-encoders were trained for one epoch.

We used silhouette metric to evaluate cluster consistency, as the popular elbow method is not ideal to fit the clusters in production because it needs a person’s input. The system then uses the highest silhouette obtained after clustering.

\section{Results}

The first set of experiments was to validate the forecaster net’s architecture. The baseline to beat was an ARIMA model for each political subdivision (33 in total) trained on the whole history to predict a 7-day window. 
A single net was trained to predict all 33 series with windows of size of 10 and a forecast window of size seven. As seven is the total size of the test set, the net does not use any test data to predict the forecast, hence avoiding data leakage.  Another neural net was trained without the day embeddings in order to validate the hypothesis that this feature helps to overcome series with peaks due to delayed tests. The seven days chosen for validation was the week from August 23 to August 29. This time window seemed appropriate, as it was during this week that the pandemic reached the peak of new daily infected cases in Colombia.
 The number of epochs used was four and the optimizer chosen was ADAM with a learning rate of $0.001$. The training data was the daily number of infected people from March 3 until August 22.

The results are summarized in Table \ref{table2}, the chosen metric to evaluate the forecast was SMAPE, rather than MAPE as there were a lot of series with a value of 0 in several days and hence the MAPE could not be computed on those cases. The SMAPE reported is the average SMAPE for each of the seven days forecast for the 33 Colombian departments.

\begin{table}[!htbp]
\caption{forecast experiments results}
\label{table2}

 \centering
\scalebox{0.9}{
\begin{tabular}{l|l|l}
\toprule
model                                                                     & Average SMAPE   & \begin{tabular}[c]{@{}l@{}} A. SMAPE in\\ biggest departments(4) \end{tabular} \\ \midrule
base-line                                                                 & 0.5769          & 0.3557                                                                     \\ 
Neural net                                                                & 0.5932          & \textbf{0.3437}                                                            \\ 
\begin{tabular}[c]{@{}l@{}}Neural net with \\ day-embeddings\end{tabular} & \textbf{0.5717} & 0.3861    
\\ \bottomrule
\end{tabular}}
\end{table}

It is worth noticing that even though the neural net with the day embeddings had a better overall performance, it actually did worse on average in Colombia's Departments with the highest population (Bogotá, Antioquia, Valle del Cauca and Atlántico ).

The second set of experiments were relative to the clustering process. Once the new cases of infection were forecasted, those values were averaged and used as a feature for the data-set described in section 2. The results are summarized in Table \ref{table3}.

\begin{table}[!htbp]
\caption{Cluster experiment results}
\label{table3}
 \centering
\resizebox{\textwidth}{!}{%
\begin{tabular}{l|c c c c c c c c}
\toprule
                                   & \multicolumn{8}{c}{number of clusters}                                            \\ \toprule
experiment        & 3     & 4     & 5     & 6      & 7     & 8      & 9      & 10     \\ \midrule
No dim. reduction, k-means   & 0.243 & 0.238 & 0.225 & 0.232  & 0.190 & 0.194  & 0.162  & 0.143  \\ \hline
All Variables, k-medoids & 0.069 & 0.069 & 0.010 & -0.042 & 0.099 & 0.047  & -0.081 & -0.105 \\ \hline
PCA, k-means  & 0.247 & 0.231 & 0.230 & 0.236  & 0.195 & 0.199  & 0.168  & 0.148  \\ \hline
PCA, k-medoids & 0.071 & 0.070 & 0.012 & -0.057 & 0.102 & -0.051 & -0.085 & -0.114 \\ \hline
Stacked Autoencoder, k-means   & 0.294 & 0.338 & 0.372 & 0.354  & 0.323 & 0.326  & 0.335  & 0.329  \\ \hline
Stacked Autoencoder, k-medoids & 0.249 & 0.268 & 0.216 & 0.195  & 0.111 & 0.202  & 0.190  & 0.303  \\ \hline
Shared Weights Autoencoder, k-means   & 0.353 & 0.408 & \textbf{0.426} & 0.422  & 0.416 & 0.373  & 0.370  & 0.358  \\ \hline
Shared Weights Autoencoder, k-medoids & 0.274 & 0.321 & 0.388 & 0.319  & 0.259 & 0.235  & 0.216  & 0.196  \\ \hline
GA - All Variables, k-means  & 0.412 & 0.386 & \textbf{0.414} & 0.385  & 0.396 & 0.369  & 0.349  & 0.333  \\ \hline
GA - All Variables, k-medoids & 0.335 & 0.303 & 0.259 & 0.241  & 0.261 & 0.193  & 0.219  & 0.208  \\ \bottomrule
\end{tabular}
}
\end{table}

\begin{figure}[!htbp]
  \centering
  \subfloat[GA, number of clusters: 5]{\includegraphics[scale=0.1]{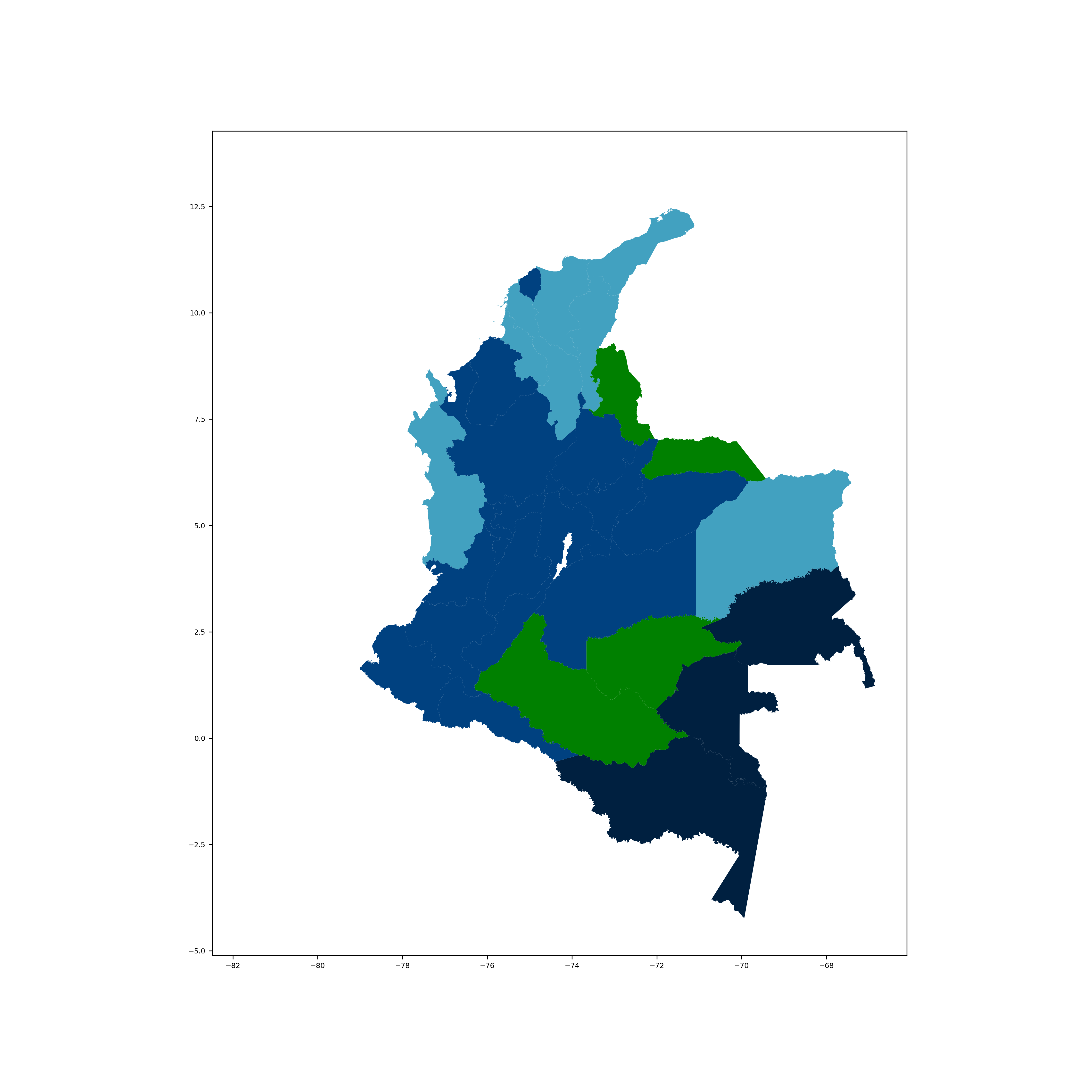}\label{f3}}
  \hspace{1em}
  \subfloat[Symmetric auto-encoder, number of clusters: 5.]{\includegraphics[scale=0.1]{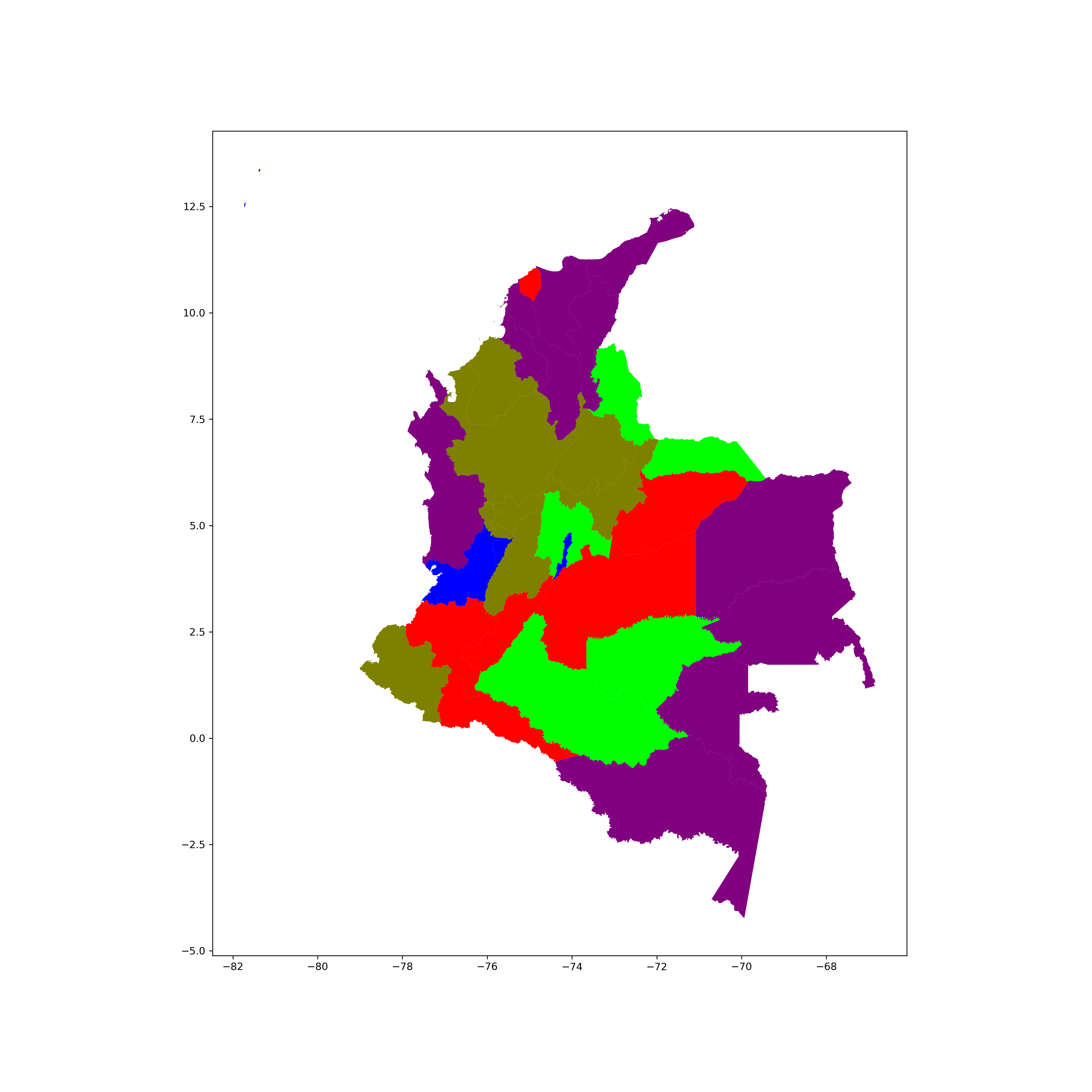}\label{f4}}
  \caption{Clusters with the highest silhouette obtained.}
\label{fig:cluster}
\end{figure}

For the case of the symmetric auto-encoder and the genetic algorithms, the suggested number of clusters was very satisfactory (Figure \ref{fig:cluster}). The clusters obtained from the symmetric auto encoder embedings clearly distinguished the four most densely populated Colombian departments. In contrast, the genetic algorithms produced one big contiguous cluster where only the capital city (Bogota) is differentiated.

\section{Conclusions}
Machine learning methods could provide reliable systems to assist in the development of public policies. In this work we proposed and tested a simple approach to generate groups of regions in Colombia with similar social, economic and health impact by the Covid-19 pandemic. The results obtained were satisfactory, the system naturally suggests
five clusters and distinguishes departments with high populations regions from their surroundings. 

Important questions arise for future works, such as the analysis of cluster centroids to gain some interpretability on the clustering process. Another downstream task is to compare the clusters generated at different dates by the system, and try to identify patterns related to vulnerable population, or the economic impact of a lockdown. We hope that this work will encourage more applications of machine learning related to the development of public policies.
\nocite{web:lang:stats}

\begin{ack}
We want to thank professor José Aguilar for supporting the project and the comments he gave on the manuscript also, to researcher Nicolás Páez for the fruitful discussions. Finally to EAFIT university and Colombian government that funded this work.
\end{ack}

\printbibliography

@article{smyl2020hybrid,
  title={A hybrid method of exponential smoothing and recurrent neural networks for time series forecasting},
  author={Smyl, Slawek},
  journal={International Journal of Forecasting},
  volume={36},
  number={1},
  pages={75--85},
  year={2020},
  publisher={Elsevier}
}

@inproceedings{sutskever2014sequence,
  title={Sequence to sequence learning with neural networks},
  author={Sutskever, Ilya and Vinyals, Oriol and Le, Quoc V},
  booktitle={Advances in neural information processing systems},
  pages={3104--3112},
  year={2014}
}

@article{oord2016wavenet,
  title={Wavenet: A generative model for raw audio},
  author={Oord, Aaron van den and Dieleman, Sander and Zen, Heiga and Simonyan, Karen and Vinyals, Oriol and Graves, Alex and Kalchbrenner, Nal and Senior, Andrew and Kavukcuoglu, Koray},
  journal={arXiv preprint arXiv:1609.03499},
  year={2016}
}

@article{lau2020positive,
  title={The positive impact of lockdown in Wuhan on containing the COVID-19 outbreak in China},
  author={Lau, Hien and Khosrawipour, Veria and Kocbach, Piotr and Mikolajczyk, Agata and Schubert, Justyna and Bania, Jacek and Khosrawipour, Tanja},
  journal={Journal of travel medicine},
  volume={27},
  number={3},
  pages={taaa037},
  year={2020},
  publisher={Oxford University Press}
}

@article{feng2020rational,
  title={Rational use of face masks in the COVID-19 pandemic},
  author={Feng, Shuo and Shen, Chen and Xia, Nan and Song, Wei and Fan, Mengzhen and Cowling, Benjamin J},
  journal={The Lancet Respiratory Medicine},
  volume={8},
  number={5},
  pages={434--436},
  year={2020},
  publisher={Elsevier}
}

@article{dell2020economic,
  title={Economic policies for the COVID-19 war},
  author={Dell’Ariccia, Giovanni and Mauro, Paolo and Spilimbergo, Antonio and Zettelmeyer, Jeromin},
  journal={IMF Blog},
  volume={1},
  year={2020}
}

@book{friedman2001elements,
  title={The elements of statistical learning},
  author={Friedman, Jerome and Hastie, Trevor and Tibshirani, Robert},
  volume={1},
  number={10},
  year={2001},
  publisher={Springer series in statistics New York}
}

@article{lleti2004selecting,
  title={Selecting variables for k-means cluster analysis by using a genetic algorithm that optimises the silhouettes},
  author={Llet{\i}, Rosa and Ortiz, M Cruz and Sarabia, Luis A and S{\'a}nchez, M Sagrario},
  journal={Analytica Chimica Acta},
  volume={515},
  number={1},
  pages={87--100},
  year={2004},
  publisher={Elsevier}
}

@article{jin2015deep,
  title={Deep learning with s-shaped rectified linear activation units},
  author={Jin, Xiaojie and Xu, Chunyan and Feng, Jiashi and Wei, Yunchao and Xiong, Junjun and Yan, Shuicheng},
  journal={arXiv preprint arXiv:1512.07030},
  year={2015}
}

@article{carrillo2020using,
  title = {{Using country-level variables to classify countries according to the number of confirmed COVID-19 cases: An unsupervised machine learning approach}},
  author={Carrillo-Larco, Rodrigo M and Castillo-Cara, Manuel},
  journal={Wellcome Open Research},
  volume={5},
  number={56},
  pages={56},
  year={2020},
  publisher={F1000 Research Limited}
}

@article{rahman2020data,
  title = {{Data-driven dynamic clustering framework for mitigating the adverse economic impact of Covid-19 lockdown practices}},
  author={Rahman, Md Arafatur and Zaman, Nafees and Asyhari, A Taufiq and Al-Turjman, Fadi and Bhuiyan, Md Zakirul Alam and Zolkipli, MF},
  journal={Sustainable Cities and Society},
  volume={62},
  pages={102372},
  year={2020},
  publisher={Elsevier}
}

@misc{web:lang:stats,
  author = {Camila, Franco and Luis, Jaramillo and Nicolas, Paez},
  title = {Estudio de la Geografía Sanitaria de Colombia (Study of the Sanitary Geography of Colombia)},
  year = {2013},
  note = {Technical Report, Ministerio de Salud y Proteccion Social (Spanish)},
  url = {https://www.minsalud.gov.co/sites/rid/Paginas/freesearchresults.aspx?k=paez}
}

\end{document}